# Learning Texture Transformer Network for Light Field Super-Resolution

Javeria Shabbir

M. Zeshan Alam

M. Umair Mukati

College of Computing, Georgia Institute of Technology

Department of Computer Science, Brandon University

GN Jabra A/S

Contrary to the traditional camera, the light field (LF) camera captures light rays approaching its surface, preserving the angular information of the incident light rays on the camera's sensor. Light field acquisition can be done in a variety of ways, including micro-lens arrays (MLAs) [1], coded masks [2], and camera array [5]. Among these different implementations, MLA based LF cameras offer a cost-effective approach. However, there lies a spatio-angular tradeoff in this design, since a single sensor is shared to capture both spatial and angular information.

To overcome this tradeoff, recently, some learning-based methods have been proposed [3, 6, 7]. In [7], a residual convolutional network is utilized to achieve high spatial resolution. Whereas, in [3] each sub-aperture image is super resolved individually and then to maintain parallax structure, a regularization network was appended. In [6], a texture transformer network is proposed that transfers high-quality texture from the reference image for target image generation. Motivated by this design, we propose to utilize a high-quality reference image to improve the resolution of all the views of the light field.

Our target is to improve the spatial resolution of light-field images by four times. We propose a modular technique comprising three different modules to achieve this task. The first module, named All-In-Focus High-Quality Reference Generator (AHQRG), generates a high-resolution image of the central view of the input low-resolution LF. The second module is the Texture Transformer Network for Image Super-Resolution (TTSR), which takes in two inputs: the output of the AHQRG module, which is treated as a reference image, and the low-resolution perspective image of the light field. TTSR is used to generate high-resolution views of the LF sequentially. However, since each view of the LF is super-resolved independently, the entire LF may not follow the regularity in the LF structure. This knowledge is taken as prior to further improve the LF's spatial resolution in the third module called LF refinement (LFREFINE).

*All-In-Focus High Quality Reference Generator*: We adopted spatial and angular interleaved convolution from [3] and modified this network such that it takes 7×7 views of the LF as input and results in a single view at the output. AHQRG module shown in Figure 1, has 3×3 convolutional layers followed by a set of four interleaved filters that alternates between the angular and spatial representation of LFs. Three 3d convolution layers are introduced after the interleaved filters to extract information from the spatial and angular coordinates simultaneously. Finally, the resulting residual is added to the 4× bicubically upsampled central view producing a high-quality central view.

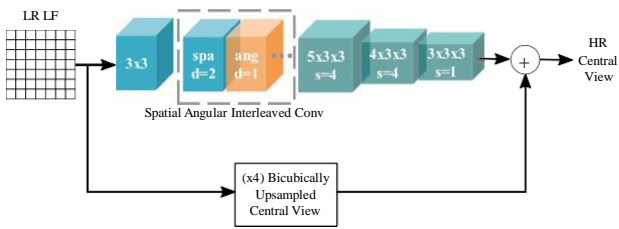

Figure 1: Block diagram of All-In-Focus High Quality Reference Generator (AHQRG).

*Texture Transformer Network:* TTSR module sequentially takes one low-resolution LF view and a high-resolution reference image produced by AHQRG module and outputs a corresponding high-resolution view of the LF. TTSR extracts feature from the reference image using a texture transformer to super-resolve low-resolution image. Bicubically down-sampled and up-sampled reference image serves as key for texture transformer, bicubically up-scaled low-resolution image as query and reference image as value. Texture transformer successfully transfers high-resolution features from reference to low-resolution image. The performance is further enhanced as the features at different scales produced by multiple texture transformers are combined to create a high-resolution image.

*Light-field Refinement:* Since each view of the LF is super-resolved using TTSR independently, the resulting LF may not follow the LF structure. We take this problem as an opportunity to further enhance the quality of the LF by imposing the LF constraint. Similar research is done in [3] to remove the artifacts during the view synthesis process. We utilize the LF refinement step (known as LF blending module) to improve the overall quality. Furthermore, EPI loss function is incorporated to enforce the LF prior.

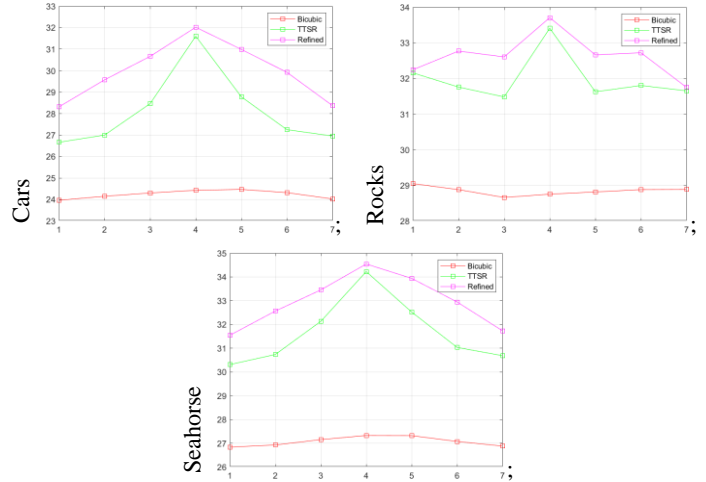

Figure 2: Comparison of the proposed method with bicubically resized view, and outputs from TTSR three different LFs from [4] dataset

*Performance Evaluation:* We quantitatively compare the performance of the proposed method with the bicubically resized LF and the output of sequentially super-resolved LF views from TTSR. We utilize PSNR as an evaluation metric. For simplicity of representation, we plotted the PSNR of only the diagonal views of the LF images in Figure 2. A significant gain in PSNR is evident for the proposed method as compared to the bicubically resized views. Though TTSR can generate a high-quality central view, as soon as the view deviates from the central location, the quality starts dropping. We believe this may be due to incorrect texture placement. On the other hand, LFREFINE considerably improves performance by applying LF prior. This module improves the quality of the views away from the central location as well as the central view.

[1] M. Z. Alam and B. K. Gunturk. Hybrid light field imaging for improved spatial resolution and depth range. *Machine Vision and Applications*, 29:11–22, 2018.

[2] M. Z. Alam and B. K. Gunturk. Deconvolution based light field extraction from a single image capture. In *IEEE Intl. Conf. on Image Processing*, pages 420–424, 2018.

[3] J. Jin, J. Hou, H. Yuan, and S. Kwong. Learning light field angular super-resolution via a geometry-aware network. In *AAAI Conf. on Artificial Intelligence*, volume 34, pages 11141–11148, 2020.

[4] N. K. Kalantari, T. Wang, and R. Ramamoorthi. Learning-based view synthesis for light field cameras. *ACM Trans. on Graphics*, 2016.

[5] B. Wilburn, N. Joshi, V. Vaish, E. V. Talvala, E. Antunez, A. Barth, A. Adams, M. Horowitz, and M. Levoy. High performance imaging using large camera arrays. *ACM Trans. on Graphics*, pages 765–776, 2005.

[6] F Yang, H Yang, J Fu, H Lu, and B Guo. Learning texture transformer network for image super-resolution. In *IEEE Conf. on Computer Vision and Pattern Recognition*, pages 5791–5800, 2020.

[7] S. Zhang, Y. Lin, and H. Sheng. Residual networks for light field image super-resolution. In *IEEE Conf. on Computer Vision and Pattern Recognition*, pages 11046–11055, 2019.